\documentclass{isprs} 
\usepackage{subfigure}
\usepackage{setspace}
\usepackage{geometry} 
\usepackage{epstopdf}
\usepackage[labelsep=period]{caption}  
\usepackage[british]{babel} 
\usepackage[hang]{footmisc}
\usepackage{amsmath}
\usepackage{url}
\usepackage{booktabs}
\usepackage{algorithm}
\usepackage{algpseudocode}
\usepackage{algorithmicx}

\usepackage{todonotes}

\begin{document}

\title{CM2LoD3: Reconstructing LoD3 Building Models Using Semantic Conflict Maps}
\date{}



\author{Franz Hanke, Antonia Bieringer, Olaf Wysocki, Boris Jutzi}

 \address{
 Professorship of Photogrammetry and Remote Sensing, Technical University of Munich  
  \\ (franz.hanke, antonia.bieringer, olaf.wysocki, boris.jutzi)@tum.de\\
 }



\abstract{
Detailed 3D building models are crucial for urban planning, digital twins, and disaster management applications.
While Level of Detail 1 (LoD)1 and LoD2 building models are widely available, they lack detailed facade elements essential for advanced urban analysis. 
In contrast, LoD3 models address this limitation by incorporating facade elements such as windows, doors, and underpasses. However, their generation has traditionally required manual modeling, making large-scale adoption challenging.
In this contribution, CM2LoD3, we present a novel method for reconstructing LoD3 building models leveraging Conflict Maps (CMs) obtained from ray-to-model-prior analysis.
Unlike previous works, we concentrate on semantically segmenting real-world CMs with synthetically generated CMs from our developed Semantic Conflict Map Generator (SCMG).
We also observe that additional segmentation of textured models can be fused with CMs using confidence scores to further increase segmentation performance and thus increase 3D reconstruction accuracy.
Experimental results demonstrate the effectiveness of our CM2LoD3 method in segmenting and reconstructing building openings, 
with the 61\% performance with uncertainty-aware fusion of segmented building textures.
This research contributes to the advancement of automated LoD3 model reconstruction, paving the way for scalable and efficient 3D city modeling.
Our project is available: \textit{https://github.com/InFraHank/CM2LoD3}
%
%
}


\maketitle


\section{Introduction}\label{sec:Introduction}
\begin{sloppypar}
The detailed semantic 3D building modeling has long been a challenge in photogrammetry and computer vision. While existing methods utilizing 2D building footprints and aerial imagery enable the generation of models up to Level of Detail (LoD) 2, the robust and automatic creation of LoD3 models with detailed facade elements remains an active field of research~\cite{helmutMayerLoD3,MuellerSattlerPollefeys2019_1000098386,tang2025texture2lod3,wysocki2023scan2lod3reconstructingsemantic3d}.

Automated reconstruction of LoD3-specific facade elements is crucial for numerous applications, including urban digital twins for simulating autonomous driving functions or flood scenarios. 
Previous studies~\cite{wysockiUnderpasses} introduce Conflict Maps (CMs) that utilize laser physics and 3D building model priors to identify absent elements in building priors serving as a base to accurately delineate and reconstruct building facade elements.
However, such CMs are semantic-absent, requiring semantic segmentation stemming from additional sources such as point clouds~\cite{wysockiVisibility}, or combination of point clouds and images~\cite{wysocki2023scan2lod3reconstructingsemantic3d}. 


In CM2LoD3, we present a novel approach that facilitate automatic LoD3 model reconstruction by leveraging synthetic Semantic Conflict Maps (SCMs) to infer real-world CMs semantics.
To that end, we introduce Semantic Conflict Map Generator (SCMG) that utilizes procedurally generated LoD3 models and ubiquity of facade image benchmarks to generate diverse distribution of training samples.
In presence of textured buildings, we present an uncertainty-aware fusion strategy enabling combination of SCMs with semantic segmentation of textures.
This fusion enables more robust detection of facade elements, particularly in challenging scenarios involving transparent surfaces, occlusions, or multiple layers within the facade.

\begin{figure}[th!]
\begin{center}
 	\includegraphics[width=\linewidth]{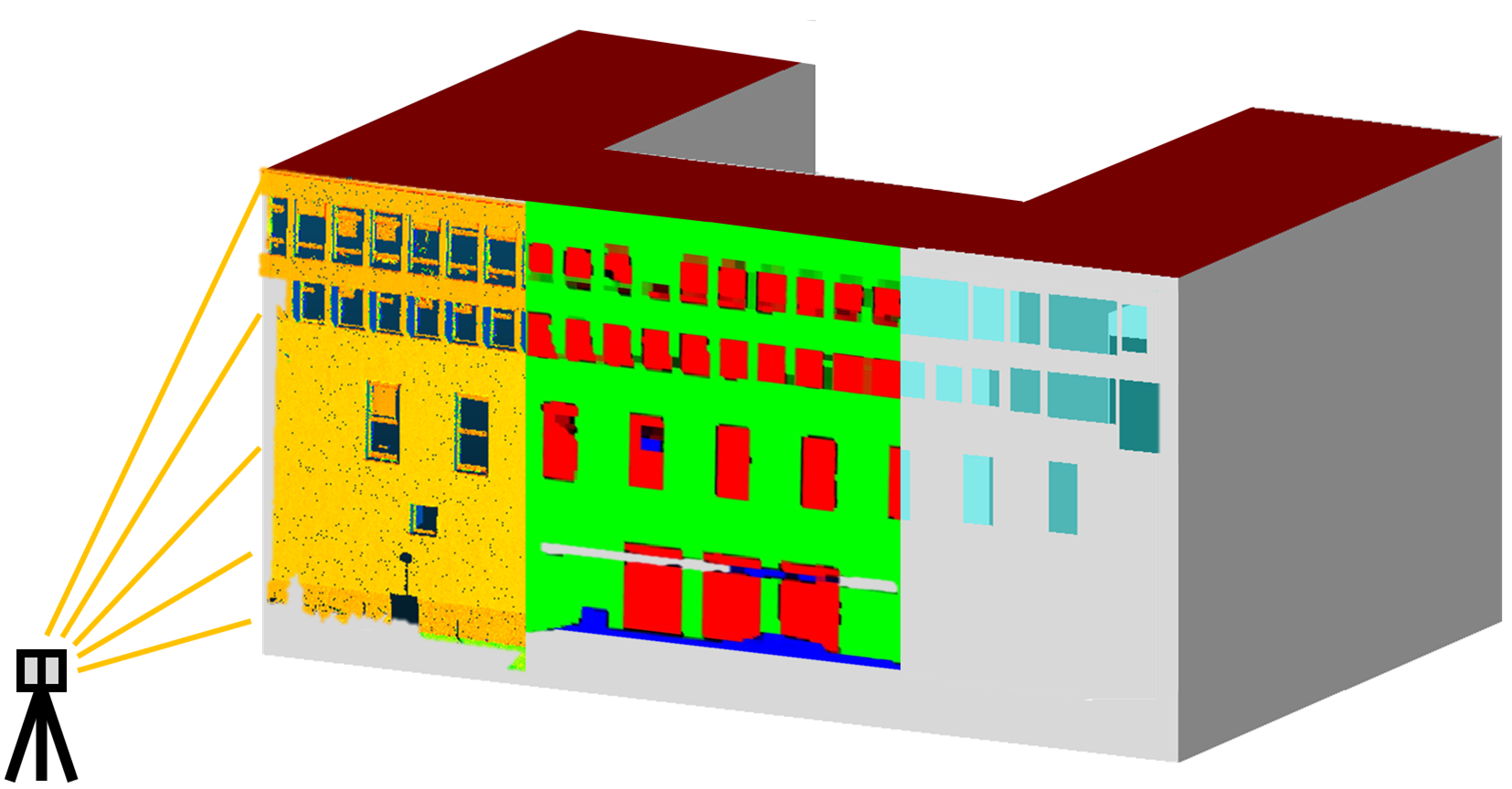}
 	\caption{CM2LoD3 leverages ray-to-model-prior analysis to obtain Conflict Maps (CMs) (left), the Semantic Conflict Map Generator (SCMG) enables inferring their semantics (center), which in turn allows for  LoD3 reconstruction (right).} 
\label{fig:Teaser}
\end{center}
\end{figure}

Our main contributions are as follows:
\begin{itemize}
    \item We introduce the CM2LoD3 method leveraging synthetic Conflict Maps (CMs) to infer real-world Semantic Conflict Maps (SCMs) for high-detail LoD3 reconstruction.
    \item We present the Semantic Conflict Map Generator (SCMG) that enables generating training data for inferring SCMs from synthetic models and facade image benchmarks.
    \item We design and develop an uncertainty-aware fusion strategy for combining SCMs and image-based facade segmentation for  robust LoD3 reconstruction.  
\end{itemize}








\end{sloppypar}
\section{Related Work}\label{sec:RelatedWork}
\begin{sloppypar}
We devote our work to the reconstruction of LoD3 building models. 
Therefore, in this Section, we present the state-of-the-art in semantic 3D city models and LoD3 reconstruction. 



\begin{figure*}[]
\begin{center}
		\includegraphics[width=\linewidth]{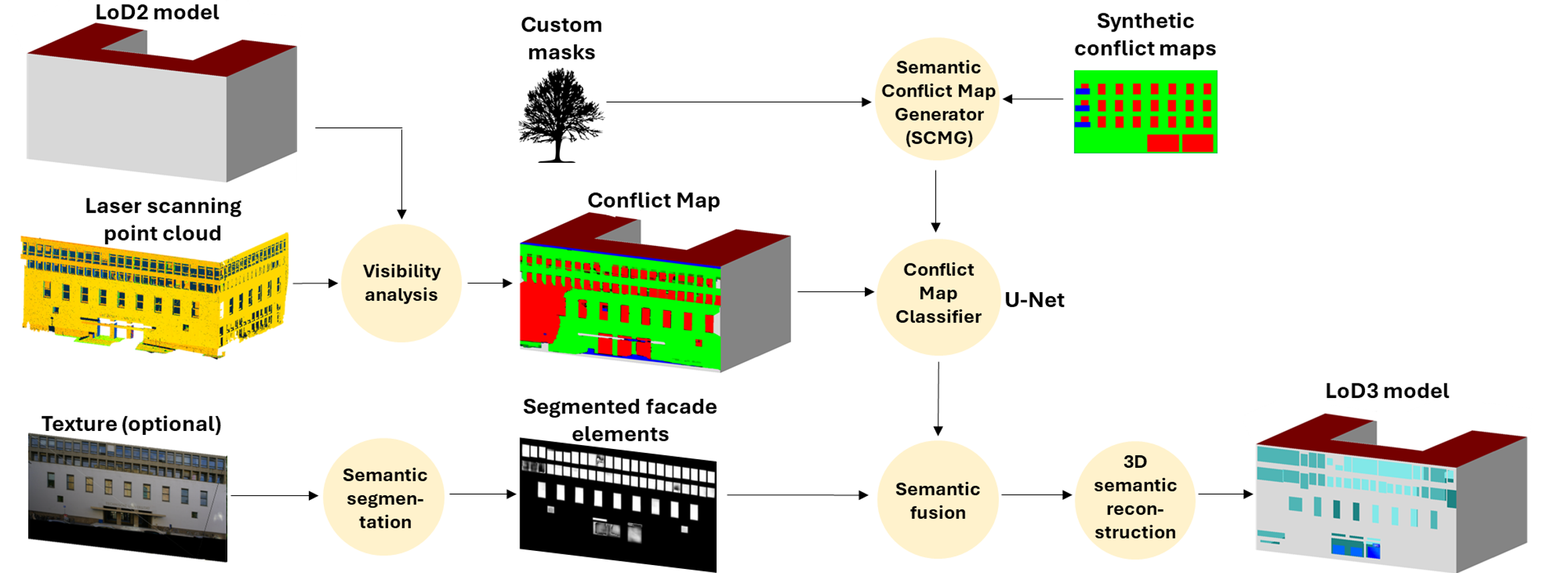}
\caption{Overview of our CM2LoD3 method for reconstructing LoD3 building models using Semantic Conflict Maps (SCMs). The process commences with ray-to-prior visibility analysis, the obtained CMs are inferred based on our synthetically generated SCMs in Semantic Conflict Map Generator (SCMG), and inferred in the adapted U-Net classifier, the uncertainty-aware fusion with image segmentation input allows for robust LoD3 reconstruction. }
 
\label{fig:Workflow}
\end{center}
\end{figure*}

\subsection{Semantic 3D City Model Standard}

Semantic 3D city models serve as comprehensive digital representations of urban structures, classifications, and spatial relationships across local, regional, and national scales. To be able to describe these models in a standardized way, the CityGML standard has been developed. 
It is an internationally recognized standard established by the Open Geospatial Consortium (OGC) \cite{grogerOGCCityGeography2012}. CityGML data can be encoded using either Geography Markup Language (GML) or CityJSON \cite{Kutzner2020,ledoux2019cityjson}.

Buildings play a fundamental role in shaping Urban Digital Twins and describing as well as documenting cities \cite{biljeckiApplications3DCity2015}. By defining urban elements in terms of 3D geometry, appearance, topology, and semantics, the current CityGML standard supports three Levels of Detail (LoD) \cite{Kutzner2020}. 
The most commonly used building representations - LoD1 and LoD2 - are extensively available, with more than 215 million open source models in use across Germany, Japan, the Netherlands, Switzerland, the United States, and Poland \cite{wysocki2024reviewing}.

Unlike conventional mesh models, semantic 3D building models offer several advantages. They are georeferenced and enriched with object-level geometric and semantic information. Additionally, they follow a hierarchical data structure that encodes inter-object relationships. These models are typically watertight and low-poly, allowing for accurate volumetric interpretation. This is achieved by integrating externally observable surfaces into a boundary representation (B-Rep) format \cite{Kolbe2021,grogerOGCCityGeography2012}.

The lowest level building description is LoD1. Such models provide basic extruded volumes based on building footprints and height data. 
LoD2 models display more complex roof geometry and can be automatically derived using open source and proprietary software, provided that aerial observation and building footprints are available \cite{RoschlaubBatscheider,HAALA2010570,munumer2022exploring}. 
%
Despite their advantages, LoD1 and LoD2 models lack detailed facade elements. LoD3 models address this limitation by incorporating facade elements such as windows, doors, balconies, and underpasses \cite{grogerOGCCityGeography2012,wysocki2024reviewing}. 


\subsection{LoD3 Reconstruction}
Reconstructing semantic 3D buildings at Level of Detail 3 (LoD3) remains a significant challenge in the fields of photogrammetry and computer vision. 
However, the generation of these high-resolution elements has traditionally relied on manual modeling \cite{manualLoD3seismic}. Efforts to automate LoD3 reconstruction have been a major focus of recent research, driving progress in its practical implementation. 

Despite recent progress, LoD3 models are still relatively rare in practice \cite{wysocki2023scan2lod3reconstructingsemantic3d,pantoja2022generating,pang20223d,KadaFacades,wang2024framework,salehitangrizi20243d}. One of the key challenges is the limited robustness of existing reconstruction methods when applied at scale. 
Many approaches rely on controlled data acquisition setups, requiring precise co-registration of multiple images and point clouds; and full unobstructed coverage of individual buildings.
This often entails capturing isolated structures such as standalone houses using 360-degree drone flights \cite{pantoja2022generating,helmutMayerLoD3}, which poses significant limitations for broader applicability.

The concept of Conflict Maps (CMs) for LoD3 reconstruction is introduced by recent works \cite{wysockiUnderpasses}. 
The CM is a surface texture obtained in the visibility analysis process: The \textit{confirmation} state is assigned when a laser ray point lies within the tolerance of prior low-level building surface; The \textit{conflict} is assigned  when a laser ray penetrates the surface; and is \textit{unknown} when otherwise (e.g., occlusions, unmeasured space).
The CMs has been been further developed to account for uncertainty-aware CMs utilizing Bayesian networks when dealing with combination of CMs and segmented point clouds \cite{wysockiVisibility}; CMs, segmented point clouds, and segmented images \cite{wysocki2023scan2lod3reconstructingsemantic3d}.
The incompleteness of CMs is also addressed by introducing an inpainting technique \cite{froech2025facadiffy}.
However, such CMs are semantic-absent, requiring semantics acquired from from additional sources such as point clouds or images.

\end{sloppypar}

\section{Methodology}\label{sec:Methodology}
\begin{sloppypar}




Our methodology for reconstructing LoD3 building models using semantic Conflict Maps (CMs) uses different steps as illustrated in Figure \ref{fig:Workflow}. Our process starts with visibility analysis in Subsection \ref{sec:visAn}. In this section, we produce CMs out of laser scanning point clouds. In Subsection \ref{sec:ConflictMapGenerator}, we generate more training data by introducing the Conflict Map Generator (CMG). The data generated in these two subsections is then used in Subsection \ref{sec:ConflictMapClassifier} to create a classifier for CM classification. Optionally, the results of this classifier can be fused with image segmentation results, described in Subsection \ref{sec:Fusion}. From the final classification, we derive a semantic 3D city model, see Subsection \ref{sec:reconstruction}. Our project is available: \textit{https://github.com/InFraHank/CM2LoD3}







\subsection{Visibility Analysis}
\label{sec:visAn}


To be able to also input real world data to our network to achive more promising results in real world scenarios, we used the workflow of developed by recent work \cite{froech2025facadiffy} to generate CMs out of laser scanning data. We assume stationary measurement at given timestamps.
The rays $\mathbf{r_i}$ are then refined as follows:

\begin{equation}
    \mathbf{r_i} = \mathbf{v} + \frac{\mathbf{p_i}-\mathbf{v}}{\mid\mathbf{p_i}-\mathbf{v}\mid}
\end{equation}

In this equation, $\mathbf{v}$ describes the viewpoint and $\mathbf{p_i}$ stands for each laser scanning point. One ray is created for each laser scanning point. The ray is then originating from the viewpoint and pointed towards the laser scanning point \cite{froech2025facadiffy}.

\begin{figure}[h!]
\begin{center}
		\includegraphics[width=0.75\linewidth]{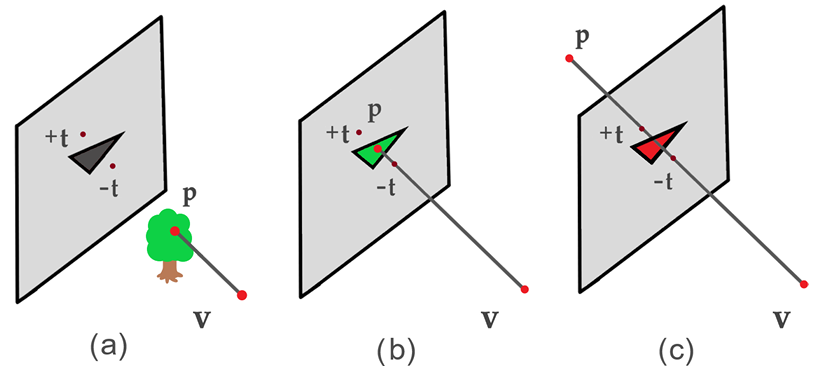}
	\caption{Principle of the Conflict Map (CM) generation \protect\cite{froech2025facadiffy}. a) unknown (grey) b) confirmed (green) c) conflict (red)}
\label{fig:conflictmapgen}
\end{center}
\end{figure}
Figure \ref{fig:conflictmapgen} depicts a visual explanation. 
First, ray casting is performed on laser scans and LoD2 models. Then the distance of each laser scanning point is compared to the corresponding LoD2 model (depending on the viewpoint). If the point is laying behind the model, then the point is marked as \textit{conflict} and has the potential to be a building opening. The corresponding pixel in the resulting image is then colored in red. If the point lays in front of the building model it is marked as \textit{unknown} and can be a disturbing tree for example. The corresponding pixel is then colored in blue. If the point is close to the facade under the assumed tolerance value $t$ then the point is marked as facade \textit{confirmed}. The pixel value then set to green. 


The tolerance has to be that high, even though the terrestrial laser scans are so precise because of the LoD2 model: The buildings we used have some bulges and therefore do not completely overlay on the whole facade with the actual point cloud. Here we can see a limitation of the LoD2 model, which is just based on the footprint of a building. With the idea of creating the CMs that way, we want to detect windows and doors as building openings. With the U-Net we further add to each building opening the semantic \textit{door} or \textit{window}.



\subsection{Semantic Conflict Map Generator}
\label{sec:ConflictMapGenerator}
%
In the absence of training data for CMs, we create the Semantic Conflict Map Generator (SCMG). 
Initially, we create a comprehensive dataset for training the classifier, comprising both synthetic and real-world data and semantics. 
For synthetic data generation, we utilized the Random3DCity module \cite{biljeckiRandom3DCity2016} to produce various building models with configurable facade arrangements. To simulate real-world imperfections and avoid overfitting, one possibility is to introduce random occlusion masks \cite{froech2025facadiffy}. Additionally, we incorporated tree silhouettes \cite{etsy} as noise elements in the images.


\begin{algorithm}[htb]
  \caption{Occlusion Masking Process}
  \label{alg:occlusion-masking}
  \begin{algorithmic}[1]
    \Require $\mathit{datasetDir},\;\mathit{randMasksDir},\;\mathit{treeMasksDir}$
    \Ensure Occluded images are saved alongside originals

    \State $\mathcal{R} \gets \Call{LoadFiles}{\mathit{randMasksDir},\,``*.png''}$
    \State $\mathcal{T} \gets \Call{LoadFiles}{\mathit{treeMasksDir},\,``*.png''}$
    \State $\mathcal{C} \gets \Call{LoadFiles}{\mathit{datasetDir}/\text{CM},\,``*.png''}$
    \State $M \gets |\mathcal{R}| + |\mathcal{T}|$
    \State $I \gets \Call{RandSample}{0\,\dots\,|\mathcal{C}|-1,\;M}$
    \State $I_r \gets \Call{RandSubset}{I,\;|\mathcal{R}|}$
    \State $I_t \gets I \setminus I_r$
    \vspace{0.2cm}

    \For{$i = 0$ to $|\mathcal{C}|-1$}
      \If{$i\in I_r$}
        \State $mask \gets \Call{NextRandMask}{}$
        \State \Call{ApplyOcclusion}{$image_i,\;mask,\;[0,0,255]$}
      \ElsIf{$i\in I_t$}
        \State $mask \gets \Call{NextTreeMask}{}$
        \State \Call{PositionMask}{$mask,\;image_i.\text{size}$}
        \State \Call{ApplyOcclusion}{$image_i,\;mask,\;[0,0,255]$}
      \Else
        \State \Call{CopyOriginal}{$image_i,\;outputPath_i$}
      \EndIf
    \EndFor
  \end{algorithmic}
\end{algorithm}

For tree positioning, we calculate the paste position $(x_p, y_p)$ based on a gaussian model that ensures realistic placement:

\begin{equation}
x_p = \mu_x + \sigma_x \cdot \mathcal{N}(0,1)
\end{equation}

where $\mu_x = \frac{W_{cm} - W_{tree}}{2}$ and $\sigma_x = \frac{W_{cm} - W_{tree}}{6}$, with constraints $0 \leq x_p \leq W_{cm} - W_{tree}$ ensuring the tree remains within image bounds horizontally. The vertical position follows:

\begin{equation}
y_p = H_{cm} + \delta - H_{tree}
\end{equation}

where $W_{cm}$ and $H_{cm}$ are the width and height of the CM, $W_{tree}$ and $H_{tree}$ are dimensions of the tree silhouette, and $\delta$ is a grounding offset parameter controlling how much the tree extends beyond the bottom edge. This approach creates a natural clustering of trees near the centerline while ensuring they appear properly grounded with a realistic vanishing point perspective.

\begin{figure}[h!]
\begin{center}
		\includegraphics[width=3in]{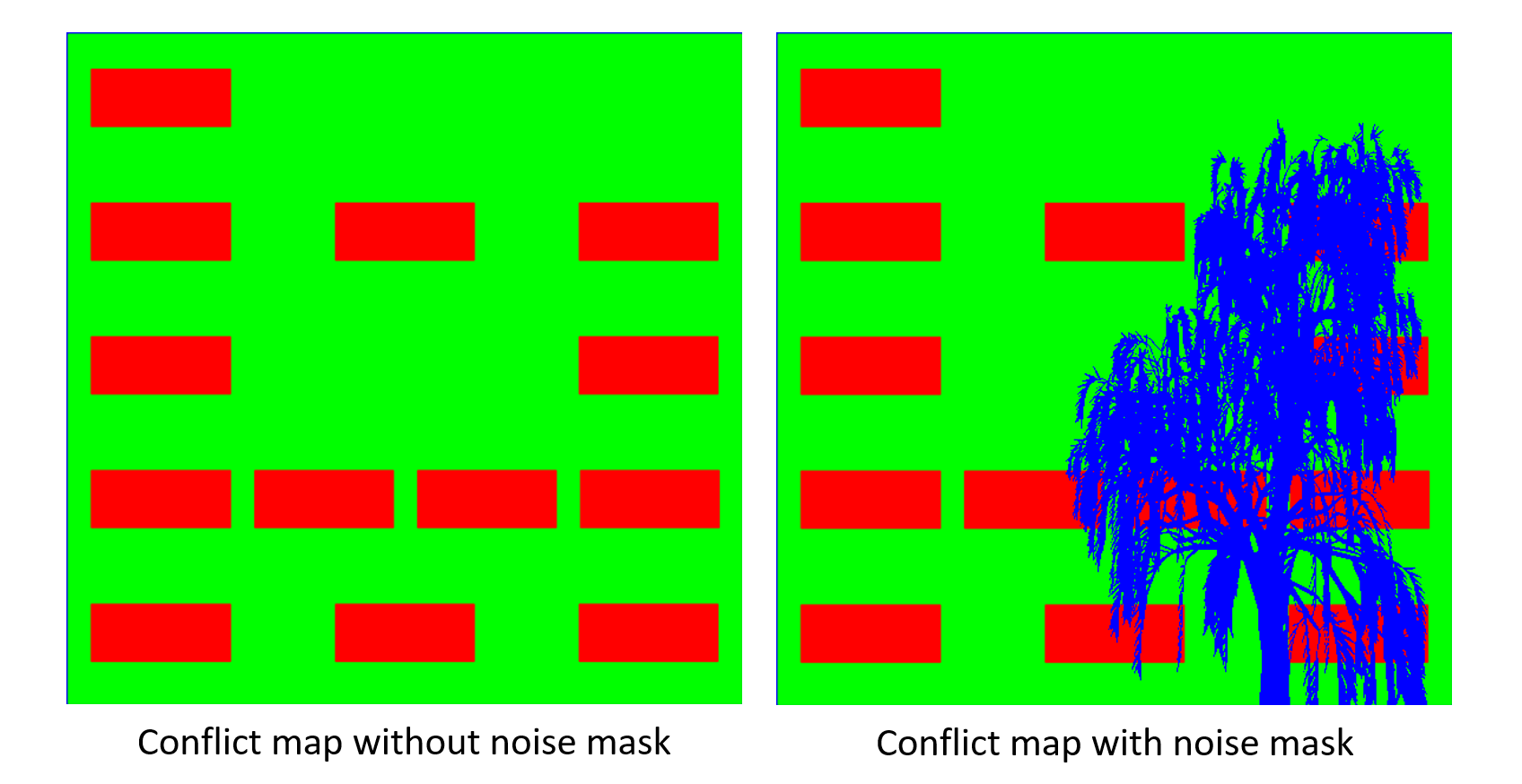}
	\caption{Semantic Conflict Map (SCM) without mask (left) and SCM with the applied tree mask (right).}
\label{fig:ConfMap_treemask}
\end{center}
\end{figure}

An example of an applied tree mask can be seen in \ref{fig:ConfMap_treemask}. Importantly, each specific mask was used only once throughout the dataset, ensuring that the network would not learn to recognize or replicate specific noise patterns, but instead develop general strategies for handling occlusions.

Additionally, we adapted the CMP facade dataset \cite{CMP-DatasetTylecek13} by mapping its diverse facade elements to our four primary classes: facade, window, door, and unknown. This mapping ensures compatibility with our classification scheme while leveraging the rich variety of architectural styles present in the dataset.


\subsection{Conflict Map Classifier} \label{sec:ConflictMapClassifier}
For CM classification, we adapted a deep learning architecture inspired by U-Net \cite{ronneberger2015unet}, which has proven effective for semantic segmentation tasks, particularly when the preservation of spatial information is critical. Our classifier processes the three-channel CMs introduced in Subsection \ref{sec:visAn} (representing confirming, unknown, and conflict states) and predicts the semantic class for each pixel. The network architecture features four encoding layers that each capture contextual information based on the different pooling outputs. The symmetric decoding path enables the precise reconstruction of an analogue to our input CM, making it particularly suitable for facade element detection.
Our UNet modification employs three key architectural changes for improved CM classification. First, we use same-padding throughout the network to preserve spatial dimensions during convolutions, ensuring the output maintains the exact 572×572 input resolution. Second, we implement size-matching interpolation with explicit dimension checking to guarantee proper feature map alignment during skip connections, resolving dimension mismatches that can occur during upsampling. Third, our network is specifically designed to process three-channel CMs, optimizing it for facade element detection rather than traditional grayscale medical imaging. These adaptations ensure pixel-perfect segmentation against our ground truth while maintaining the core strength of UNet's encoder-decoder architecture with skip connections.


Our choice of U-Net was motivated by several key factors. Firstly, maintaining transparency throughout the entire network was essential, as a U-Net with this straightforward structure enables learning of complex facade patterns without excessive architectural complexity. Feature preservation is critical since every input signal needs thorough analysis, and the skip connections (copy and crop) allow the network to maintain high-resolution features throughout, which is crucial for precisely determining element boundaries while leveraging contextual neighborhood information. Additionally, training efficiency with limited data was a significant consideration. U-Net excels in this regard, creating effective models without requiring massive amounts of annotated facade data, which is particularly valuable in specialized domains like architectural segmentation.

\subsection{Uncertainty-Aware Semantic Fusion} \label{sec:Fusion}
The results from CM classifier and from Image Semantic Segmentation in Subsection are finally fused to achieve the best possible result from both methods. For this, a metric is derived from the performance of both neural networks. Each resulting pixel is then weighted by the metric and the probability of the prediction. The weighting is done separately for door and window predictions. 

\begin{equation}
P_{door} = P_{door,U-Net} \cdot \alpha_{door} + P_{door,MaskRCNN} \cdot \beta_{door}
\end{equation}
\begin{equation}
P_{win} = P_{win,U-Net} \cdot \alpha_{win} + P_{win,MaskRCNN} \cdot \beta_{win}
\end{equation}

\noindent where:
\begin{itemize}
    \item $P_{door}, P_{win} \in [0,1]$ are the normalized final fusion probabilities for door and window classes
    \item $P_{door,U-Net}, P_{win,U-Net} \in [0,1]$ are the normalized probabilities from the U-Net operating on CMs
    \item $P_{door,MaskRCNN}, P_{win,MaskRCNN} \in [0,1]$ are the normalized probabilities from Mask R-CNN processing facade images
    \item $\alpha_{door}, \alpha_{win}, \beta_{door}, \beta_{win} \in [0,1]$ are weighting coefficients with $\alpha_{door} + \beta_{door} = 1$ and $\alpha_{win} + \beta_{win} = 1$ ensuring that the resulting probabilities remain normalized
\end{itemize}

The coefficients $\alpha$ and $\beta$ control the relative importance of each classifier's prediction for a given class. For instance, a higher $\beta_{win}$ value indicates greater confidence in the Mask R-CNN's window detection capabilities compared to the U-Net.

\begin{figure}[h!]
\begin{center}
		\includegraphics[width=\linewidth]{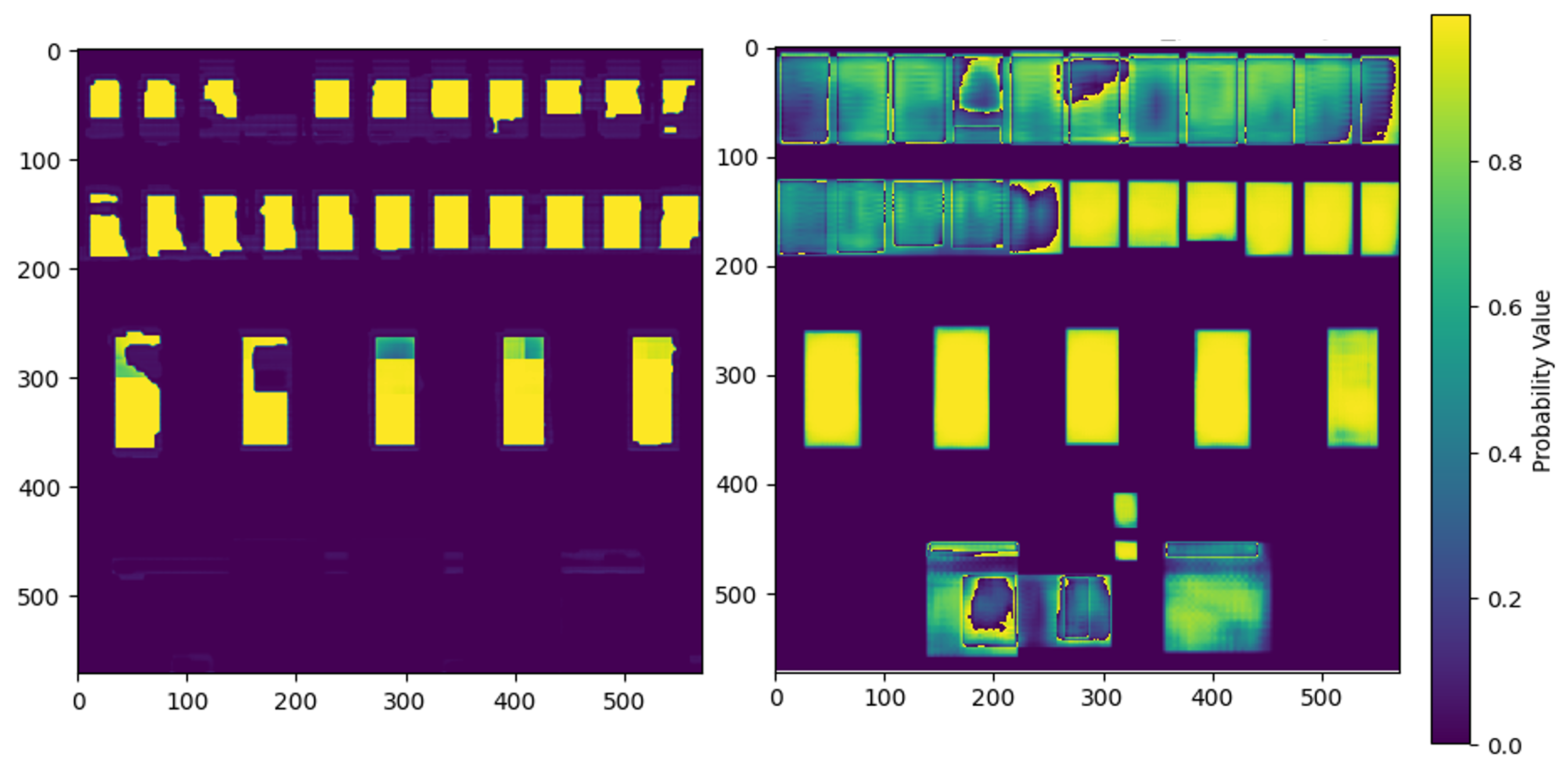}
	\caption{Probability distributions for window class: U-Net predictions from CMs (left) and Mask R-CNN predictions from facade images (right). Color intensity represents probability values from 0 to 1.}
\label{fig:U-Netprobability}
\end{center}
\end{figure}

To visualize the probability values from both networks, Figure \ref{fig:U-Netprobability} shows examples of window probability distributions. These heatmaps clearly illustrate how each network assigns confidence values across the facade, with brighter regions indicating higher probabilities of window presence.

\begin{figure}[h!]
\begin{center}
		\includegraphics[width=3in]{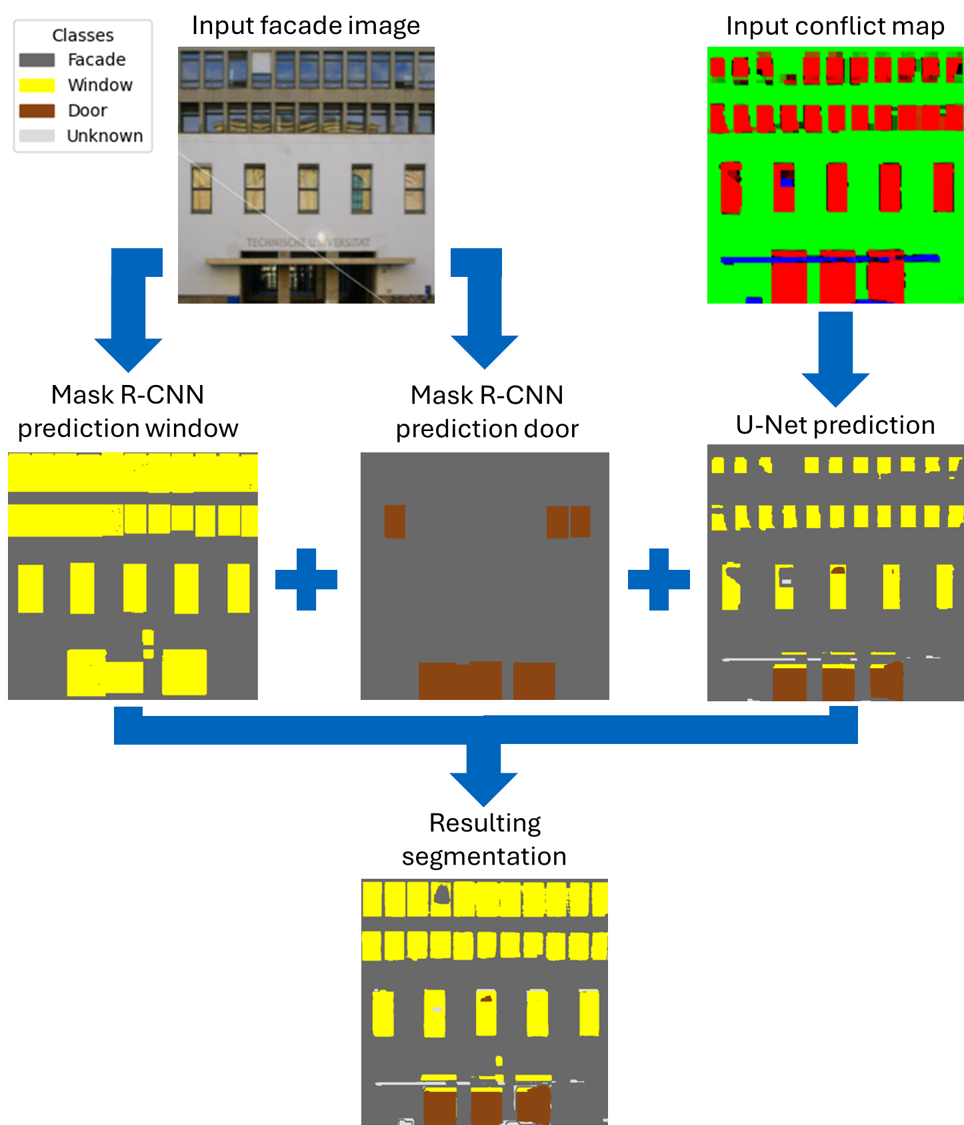}
	\caption{Example of the semantic fusion results, showing the original facade image (left), individual predictions from the U-Net on CMs and Mask R-CNN on images (middle), and the final fused segmentation (right). Gray represents facade, yellow represents windows, and brown represents doors.}
\label{fig:building57}
\end{center}
\end{figure}

Currently, we use fixed values for the weighting coefficients, which were empirically determined through extensive validation. Figure \ref{fig:building57} demonstrates the results of this semantic fusion approach applied to a sample building.

This fusion approach effectively leverages the complementary strengths of both classifiers: the U-Net excels at identifying geometric patterns in CMs while Mask R-CNN captures visual cues from facade textures. The combination produces more robust segmentation results, particularly in challenging cases where one classifier's confidence may be low. 



\subsection{Semantic 3D Reconstruction} \label{sec:reconstruction}

As final stage of our workflow we translate the fused semantic segmentation results into 3D building models using existing LoD2 building models as a foundation, similar to Wang et al. \cite{WANG202490}. This is done to ensure compliance with the CityGML standard \cite{grogerOGCCityGeography2012} since the building models are required to be solids and therefore need to be watertight. 
The integration of doors and windows into the existing model follows a two-step process: First, creating openings in the wall surfaces, and second, representing these openings as semantic objects within the CityGML hierarchy.
For each wall surface that has corresponding facade element predictions, we extract the wall's 3D geometry and establish a mapping between the 2D prediction space and the 3D wall coordinates. This mapping is crucial for accurately positioning the detected windows and doors within the 3D space. We implement this through a coordinate transformation function that 

\begin{equation}
    \text{f}(x, y) = (1-u)(1-v)P_1+ u(1-v)P_2 + (1-u)vP_3 + uvP_4
\end{equation}

where $(x,y)$ are the 2D coordinates in the prediction, $(u,v)$ are normalized coordinates, and $P_1...P_4$ are the 3D coordinates of the wall surface corners.

To represent openings correctly in the CityGML model, we implement two parallel geometric operations:

\begin{algorithm}[htb]
  \caption{Insert Openings ($ops$) into Wall Surface ($ws$)}
  \label{alg:insert-openings}
  \begin{algorithmic}[1]
    \Require $\mathit{ws},\;\mathit{ops}$
    \Ensure $ws.\text{polygon}$ has interior rings \& openings added
    \Procedure{InsertOpenings}{ws, ops}
      \For{$op \in ops$}
        \State $c \gets \Call{GetCoordinates}{op}$
        \State $r \gets \Call{ReverseWindingOrder}{c}$
        \State \Call{AddInteriorRing}{ws.polygon, r}
        \If{$op.type == \texttt{"window"}$}
          \State $w \gets \Call{CreateElement}{\texttt{"bldg:Window"}}$
          \State $w.geom \gets \Call{CreateGeometry}{c}$
          \State \Call{AddOpening}{ws, w}
        \ElsIf{$op.type == \texttt{"door"}$}
          \State $d \gets \Call{CreateElement}{\texttt{"bldg:Door"}}$
          \State $d.geom \gets \Call{CreateGeometry}{c}$
          \State \Call{AddOpening}{ws, d}
        \EndIf
      \EndFor
    \EndProcedure
  \end{algorithmic}
\end{algorithm}



This dual representation is necessary to maintain both geometric accuracy and semantic richness in the resulting models.
To ensure that the CityGML standards are still met, we implemented also some validity checkers (closed loops, no self-intersections).

\section{Experiments}\label{sec:Experiments}

\noindent \textbf{Data}
For our visibility analysis in Subsection \ref{sec:visAn}, we use terrestrial laser scanning data and facade images from the TUM2TWIN project \cite{tum2twin}. The accuracy of the laser scanning data is 0.007 m absolute and 0.001 m relative. The laser scanning data used to train the CM classifier comprised four buildings around the TUM campus. We used a total of 272 openings from those buildings. We used three building facade images for the semantic fusion. 

We used standard post-processing tasks on those images to smooth them and eliminate irrelevant small noisy blobs. We applied the morphological operation \textit{opening} on the images with a kernel size of (5, 5). By this, small holes which correspond solely to noise were removed.

The second data type to train the CM classifier is synthetic data. Those are created with the CMG, as described in Section \ref{sec:ConflictMapGenerator}.
Since those images do not contain noise and are too different from the real CMs, we manually added noise. We utilized two primary sources: 228 tree silhouettes from a publicly available dataset \cite{etsy}, and 426 random geometric masks \cite{froech2025facadiffy}.

\noindent \textbf{Parameter Settings}
The final U-Net implementation was trained with selected parameters to optimize performance on the facade segmentation task. The model was trained to classify each pixel into our four semantic categories. 

We employed a batch size of 14 and utilized the Adam optimizer with a learning rate of $2\times10^{-4}$ and weight decay coefficient of $1\times10^{-5}$. To improve convergence, we implemented a StepLR scheduler with a step size of 10 epochs and a decay factor ($\gamma$) of 0.5, effectively halving the learning rate after every 10 epochs. Early stopping was employed with a patience parameter of 8 epochs and a minimum improvement threshold ($\delta$) of $5\times10^{-5}$ to prevent overfitting.

Our network of choice converged after 41 epochs, achieving a final training loss of 0.0098 and validation loss of 0.0116. The minimal difference between training and validation loss (0.0018) indicates strong generalization capability without overfitting to the training data. This suggests that the model successfully learned to identify the distinctive features of facade elements from our CMs.


During the U-Net training, we tested different cases to optimize the results. One of them was introducing some real TLS CMs to increase the sensitivity of our U-Net to the later expected inputs. The most significant changes occurred when we included or excluded shops from the CMP dataset, respectively. For our cases, the class \textit{shops} in the CMP dataset introduces a lot of ambiguity. Therefore, we trained one U-Net while not using this class for the trained \textit{windows} areas to see the difference in the prediction. The experiments and corresponding results are documented below. 
For the tolerance value of our visibility analysis we chose empirically $t = \pm0.7m$.

\noindent \textbf{Ablation Studies Setup}
%
To evaluate the impact of real-world data on model performance, we incorporated a subset (approximately 10\%) of our collected TLS data into the training dataset alongside synthetic data from GEN and CMP sources. This approach aims to improve the model's ability to handle the complexities and noise patterns present in real-world TLS CMs that might not be adequately represented in synthetic data.

Comparing the results between the models trained solely on synthetic data (GEN + CMP) and those including real-world data (GEN + CMP + REAL), we observed several notable improvements:
The model trained with real data showed increased robustness when processing noisy CMs from real-world TLS scans. In particular, the door detection performance improved with the inclusion of real data. For the model trained on (GEN + CMP) $IoU_{door} = 0.059$. For the model trained on (GEN + CMP + REAL) $IoU_{door} = 0.142$.
As expected, exposure to real-world examples helps the network better recognize the instances of doors in the test data, which is crucial for comprehensive facade modeling.
The addition of real data also helped to reduce misclassifications between the unknown and facade classes, which suggests that the model better learned to distinguish actual building surfaces from potential scanning artifacts or obstructions.

We used Mask2Former \cite{Mask2Former} and Segformer \cite{segformer} to compare our method with other state-of-the-art image segmentation networks. Mask2Former is a universal transformer-based architecture for segmentation tasks that predicts a set of masks and their corresponding class labels using masked attention, allowing it to handle instance, semantic, and panoptic segmentation in a unified way. Instead of classifying pixels directly, it generates segmentation masks as queries and refines them iteratively using transformer decoder layers.
SegFormer, on the other hand, is a lightweight model that combines a hierarchical transformer encoder with a simple MLP decoder, focusing on efficient and accurate semantic segmentation without using heavy mask prediction mechanisms.


To further improve the robustness of our U-Net model and enhance its generalization capabilities, we introduced artificial occlusions to our generator-created CMs, drawing inspiration from the approach to facade image inpainting \cite{FritzscheInpainting2022}. These artificial masks simulate real-world occlusions that commonly affect terrestrial laser scans, such as vegetation, vehicles, and urban infrastructure.


Our experiments showed that models trained with this artificial noising approach demonstrated improved performance when processing real-world TLS CMs with actual occlusions, confirming the value of this data augmentation strategy for practical facade element detection.

As described above, we performed post-processing tasks on the CMs. Additionally, we merged the shop class to the background to avoid confusion for the U-Net.

\subsection{Results and Discussion}\label{sec:Discussion}
%
%
\begin{figure*}[h!]
 \begin{center} 		\includegraphics[width=0.8\linewidth]{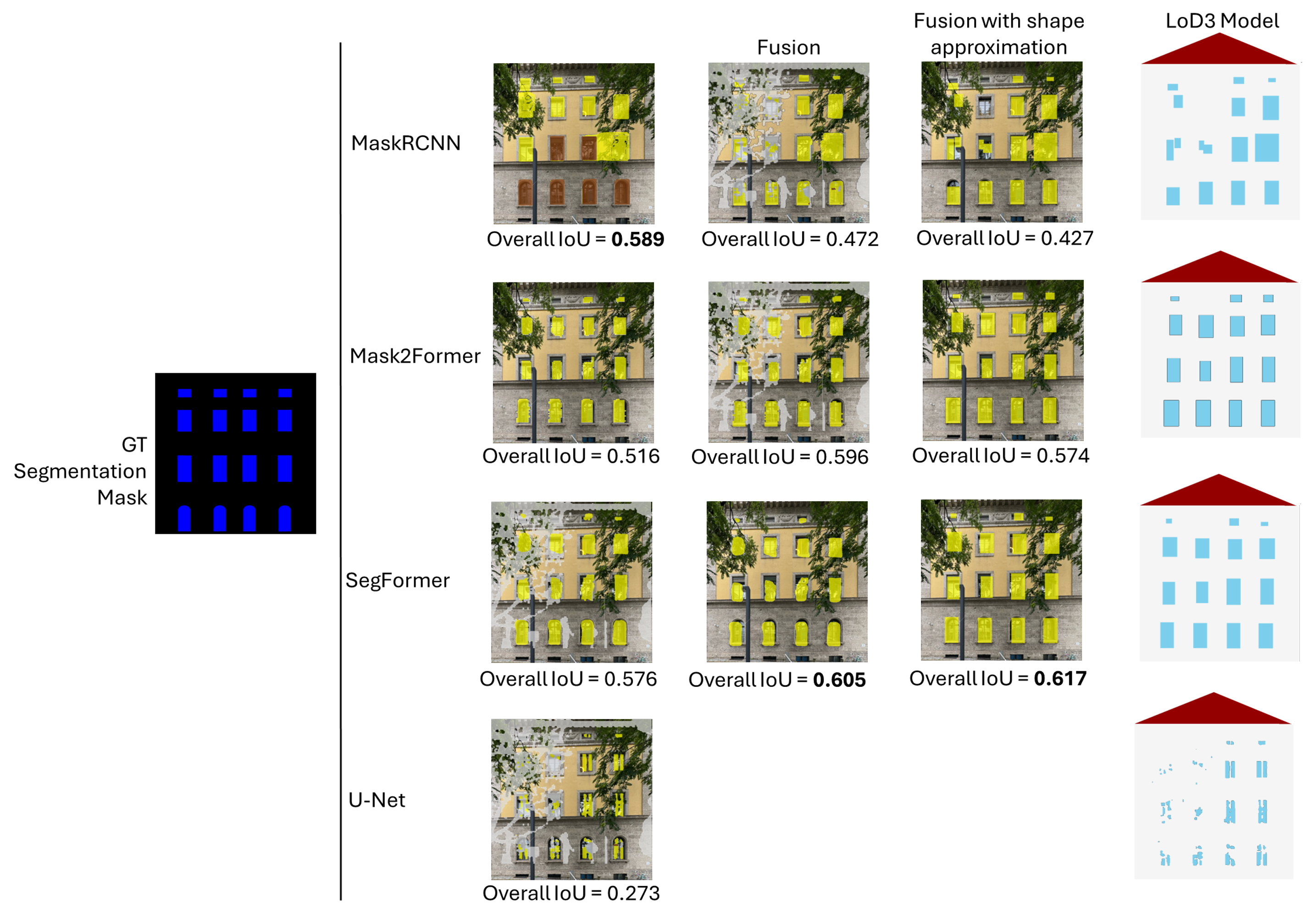}
 	\caption{Our CM2LoD3 approach tested under various ablation studies. We observe that the segmentation accuracy improves across tested baselines when fused with the U-Net trained solely on our Semantic Conflict Map Generator (SCMG). Notably, we improve robustness of Mask-RCNN in the misclassified \textit{door} class (brown) leading to skewed quantitative results owing to GT masks including only \textit{opening} labels. }
 \label{fig:Orighealed}
 \end{center}
 \end{figure*}
%
%
%
In Table \ref{tab:iou}, we show quantitative results of our CM2LoD3 method. 
The presented fusion approach outperformed single modalities, such as U-Net trained solely on Semantic Conflict Maps (SCMs) (by 0.289) and only image-based MaskRCNN (by 0.110). 
\begin{table}[h]
\centering
\caption{Comparison of segmentation approaches based on Window IoU}
\begin{tabular}{lr}
\toprule
Approach & Window IoU \\
\midrule
UNet (SCM only) & 0.178 \\
MaskRCNN & 0.357 \\
Fusion & \textbf{0.467} \\
Fusion (with shape approx.) & 0.420 \\
\bottomrule
\label{tab:iou}
\end{tabular}
\end{table}

The relatively low IoU scores observed for the U-Net can be attributed to the nature of the ground truth (GT) segmentation mask, as illustrated in Figure~\ref{fig:Orighealed}. The GT mask is relatively simplistic and does not adequately capture the qualitative improvements enabled by the contextual model. In particular, the windows are fragmented into smaller components due to the presence of window sashes, which our model detects effectively but these are not present in the GT masks. This fine-grained detection indicates that the model captures structural details well, opening new direction for future work to leverage such information for improved window style approximation.

These results suggest that the poor quantitative performance of U-Net may be misleading. The segmentation output contains not only windows and doors but also a variety of other structural elements, contributing to a more comprehensive and valuable scene representation. Despite modest numerical scores, the semantic segmentation derived from image data alone proves robust, and each evaluated method results in an overall usable output.

Furthermore, the comparison between the “Fusion” and “Fusion (with shape approximation)” approaches underscores the significance of when geometric simplification is applied. The observed performance differences reflect the alignment (or misalignment) with the expectations encoded in the ground truth. This highlights the importance of considering both quantitative metrics and qualitative outcomes in evaluating segmentation performance.

\noindent \textbf{Limitations and Future Work}
%
The quality of the CMs strongly depends on the laser scanning data. For the terrestrial laser scanner, the data is very precise and the resulting CMs are usable for our approach. Also, our training data for the U-Net are based on European buildings. Therefore, the developed workflow predicts most accurately on those type of buildings.

If we analyze the results of the U-Net (Figure \ref{fig:building57} in Subsection \ref{sec:Fusion}), we can see that the shape of the windows is very noisy. This is caused by the CMs of real data. We can not see the typical rectangular structure of the windows. In future work, when our synthetic data generation capabilities improve, we could experiment using other machine learning architectures, e.g., foundation models. 
The Mask R-CNN network also has some challenges in its predictions: For the windows, some probabilities for the class \textit{windows} are sometimes higher between the actual windows. The prediction looks blurred. There are false predictions for the doors; sometimes, windows are wrongly detected as doors. The semantic fusion should account for these problems and use the best solution of both networks to get the best possible prediction. In the image at the bottom we can see that the windows are rectangular and the doors are predicted in the right location.

\section{Conclusion}\label{sec:Conclusion}
In this paper, we propose a novel method, CM2LoD3, for improving the semantic segmentation of facade elements by leveraging automatically generated Conflict Maps (CMs) for LoD3 reconstruction.
Our approach utilizes a U-Net-based model trained with our Semantic Conflict Map Generator (SCMG) training data to improve the segmentation of facade openings. 
The experiments demonstrate that our uncertainty-aware fusion with image-based semantic segmentation models improves final LoD3 reconstruction across various tested networks.
We are convinced that conflict-based segmentation will prove essential in automatic LoD3 reconstruction due to its robustness and accuracy.
In future work, we aim to further optimize the prediction of the U-Net, compare the CM ground truth with LoD3 semantics, test different segmentation models to improve the semantic fusion accuracy, and test our approach on larger and more diverse datasets.

\noindent \textbf{Acknowledgements}
We would like to thank the team of TUM2TWIN for providing the openly accessible geodata around the TUM Campus. Especially Thomas Fröch's work in conflict map creation was a key part in realizing this project. Furthermore, we express our gratitude towards Wenzhao Tang and Weihang Li for their valuable assistance in validating our approach.
{
	\begin{spacing}{1.17}
		\normalsize
		\bibliography{bibliography} 
	\end{spacing}
}

\end{sloppypar}
\end{document}